\documentclass[pdflatex,sn-basic,iicol]{sn-jnl}
\usepackage[T1]{fontenc}

\usepackage[utf8]{inputenc}

\usepackage{multirow}
\usepackage{array}

\jyear{2023}%

\theoremstyle{thmstyleone}%
\theoremstyle{thmstyletwo}%

\theoremstyle{thmstylethree}%

\begin{document}

\newcommand{\robertuito}[0]{RoBERTuito}
\newcommand{\robertuitounc}[0]{\robertuito{}$_{uncased}$}
\newcommand{\robertuitocas}[0]{\robertuito{}$_{cased}$}
\newcommand{\robertuitodea}[0]{\robertuito{}$_{deacc}$}

\newcommand{\mbf}[1]{\mathbf{#1}}

\newcommand{\bertweet}[0]{BERTweet}
\newcommand{\bert}[0]{BERT}
\newcommand{\beto}[0]{BETO}
\newcommand{\bertin}[0]{BERTin}
\newcommand{\mbert}[0]{mBERT}
\newcommand{\roberta}[0]{RoBERTa}
\newcommand{\robertaes}[0]{RoBERTa$_{es}$}
\newcommand{\xlm}[0]{XLM-R}
\newcommand{\xlmbase}[0]{XLM-R$_{BASE}$}
\newcommand{\xlmlarge}[0]{XLM-R$_{LARGE}$}
\newcommand{\electra}[0]{ELECTRA}
\newcommand{\electricidad}[0]{ELECTRICIDAD}

\newcommand{\lince}[0]{LinCE}

\newcommand{\alberto}[0]{AlBERTo}

\newcommand{\bertit}[0]{BERT$_{it}$}

\newcommand{\mr}[2]{\multirow{#1}{*}{#2}}
\newcommand{\mc}[2]{\multicolumn{#1}{c}{#2}}

\newcommand{\pysentimiento}{\emph{pysentimiento}}
\newcommand{\tweetnlp}{\emph{TweetNLP}}
\newcommand{\stanza}{\emph{Stanza}}
\newcommand{\flair}{\emph{Flair}}
\newcommand{\textblob}{\emph{TextBlob}}
\newcommand{\vader}{\emph{VADER}}

\newcommand{\from}[3]{\textbf{[From #1 to #2]: #3}}

\title[\pysentimiento{}]{\pysentimiento{}: A Python Toolkit for Opinion Mining and Social NLP tasks}
%


\author*[1]{\fnm{Juan Manuel} \sur{Pérez}}\email{jmperez@dc.uba.ar}
\author[2,3]{\fnm{Mariela} \sur{Rajngewerc}}\email{marielaraj@gmail.com}
\author[1]{\fnm{Juan Carlos} \sur{Giudici}}\email{jgiudici@dc.uba.ar}
\author[1]{\fnm{Damián A.} \sur{Furman}}\email{dfurman@dc.uba.ar}
\author[2]{\fnm{Franco} \sur{Luque}}\email{francolq@unc.edu.ar}


\author[2]{\fnm{Laura} \sur{Alonso Alemany}}\email{lauraalonsoalemany@unc.edu.ar}
\author*[4]{\fnm{Maria Vanina} \sur{Martinez}}\email{vmartinez@iiia.csic.es}

\affil*[1]{\orgdiv{Instituto de Ciencias de la Computación}, \orgname{CONICET, Universidad de Buenos Aires}, \city{Buenos Aires}, \country{Argentina}}

\affil[2]{\orgdiv{Facultad de Astronomía, Matemática y Física}, \orgname{Universidad Nacional de Córdoba}, \city{Córdoba}, \country{Argentina}}

\affil[3]{ \orgname{CONICET}, \city{Buenos Aires}, \country{Argentina}}

\affil[4]{\orgname{IIIA-CSIC}, \orgaddress{\street{UAB Campus}, \city{Bellaterra}, \postcode{08193}, \state{Barcelona}, \country{Spain}}}


\abstract{In recent years, the extraction of opinions and information from user-generated text has attracted a lot of interest, largely due to the unprecedented volume of content in Social Media. However, social researchers face some issues in adopting cutting-edge tools for these tasks, as they are usually behind commercial APIs, unavailable for other languages than English, or very complex to use for non-experts. To address these issues, we present \pysentimiento{}, a comprehensive multilingual Python toolkit designed for opinion mining and other Social NLP tasks. This open-source library brings state-of-the-art models for Spanish, English, Italian, and Portuguese in an easy-to-use Python library, allowing researchers to leverage these techniques. We present a comprehensive assessment of performance for several pre-trained language models across a variety of tasks, languages, and datasets, including an evaluation of fairness in the results.}




\maketitle

\section{Introduction}

Extracting opinions and states of mind from user-generated context has drawn a lot of attention since the eclosion of Web 2.0 and Social Networks. Many possibilities arise for the analysis of the interactions of users on the Internet: studying the behavior of consumers \citep{horrigan2008online},  political campaigns, and even studying the changing patterns of emotions during the COVID-19 pandemics \citep{kaur2020monitoring}.

Due to the immense amount of content generated on various sites and social networks\footnote{It is estimated that 500 million tweets per day are generated worldwide by 2021. Source: \url{https://www.internetlivestats.com/twitter - statistics/}}, it is difficult ---if not impossible--- to carry out an extensive analysis of opinions without some kind of automation. As a result, lots of effort has been put into developing automatic techniques to be able to extract this type of information from texts created by users. The advancement of Natural Language Processing (NLP) techniques have made it possible to make some significant progress in this field, in spite of the particular complexities of social media interactions.

A major issue hindering more widespread usage of these opinion-mining technologies is that resources for performing these tasks are scarce. Mainly, one has to resort to paid APIs provided by companies or rely on models that are very out-of-date or even unavailable for a given language other than English. While some general-purpose NLP libraries, like SpaCy \cite{honnibal2020spacy} or Stanza \cite{qi2020stanza} have integrated sentiment analysis capabilities, however, they are not usually available for languages other than English or have a suboptimal performance for social media data.

In order to foster research integrating sentiment and other opinion-mining analysis, we present \pysentimiento{}, a Python multilingual toolkit for Social NLP tasks. \pysentimiento{} provides a simple-to-use, multilingual toolkit for opinion mining in social media. It is built on top of the \textit{HuggingFace's Transformers} library and provides a simple API for researchers to use state-of-the-art models for sentiment analysis and other social NLP tasks, with current support for Spanish, English, Italian and Portuguese. This library is released as free and open-source software\footnote{Code is available at \url{http://github.com/pysentimiento/pysentimiento}} for anyone interested in using it for research purposes.

Our contributions are the following:

\begin{itemize}
    \item We release an open-source, multilingual toolkit for opinion mining and Social NLP tasks in Python, with support for Spanish, English, Portuguese and Italian.
    \item We provide an extensive evaluation of the performance of several state-of-the-art pre-trained models for different opinion mining tasks in Spanish, English, Italian and Portuguese.
    \item We provide a small fairness assessment for Emotion Analysis in English.
    \item We compare the performance of \pysentimiento{} with other open-source tools for opinion mining.
    \item We release the best-performing models as part of our library.
\end{itemize}

\begin{figure*}[t]
    \centering
    \includegraphics[width=\textwidth]{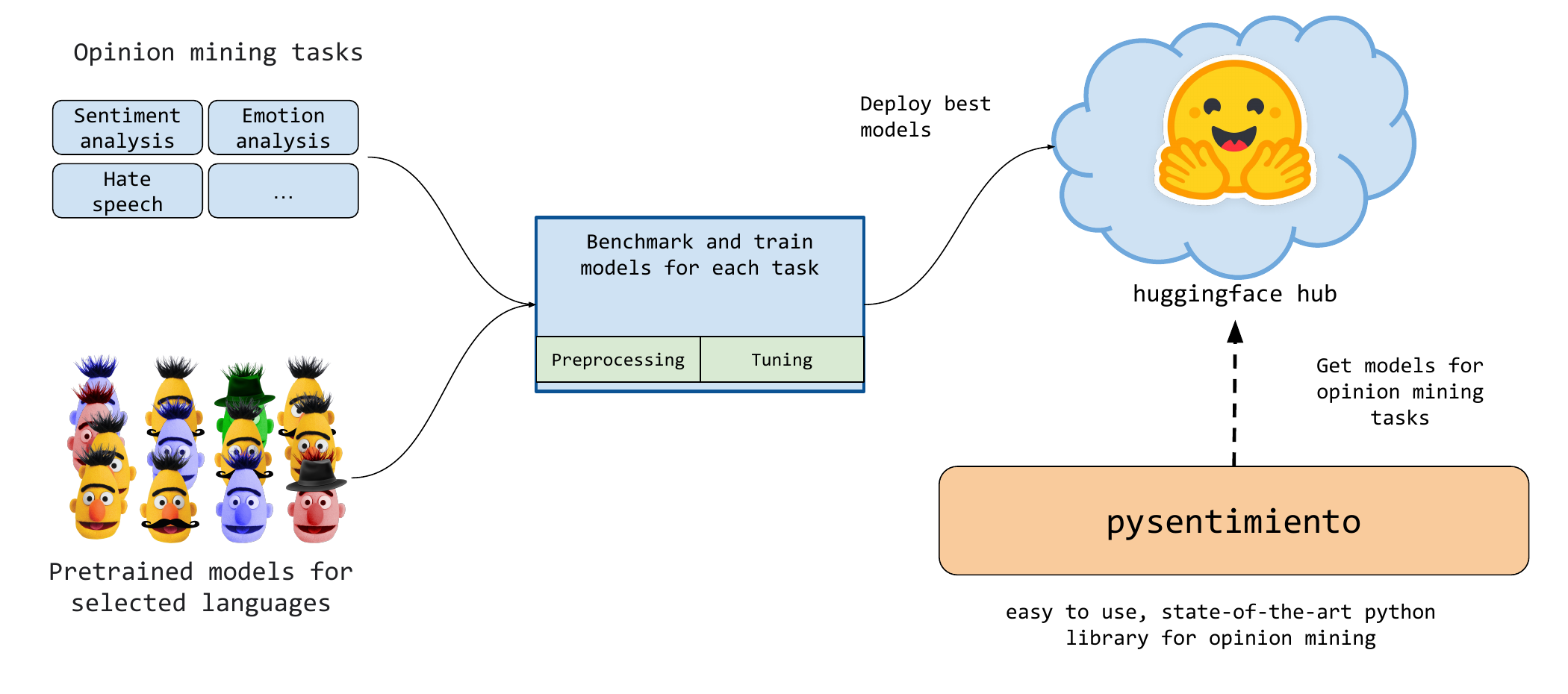}
    \caption{Description of the process presented in this paper. We selected datasets for each considered task and language pair, pre-processed them to adequate them to the format expected to train models, fine-tuned several underlying models with these datasets, compared their performance with a common benchmark, and integrated the best models in the final release of the tool and benchmarked several models on them. Selected models are deployed in the huggingface hub, and can be easily used through the library.}
    \label{fig:process_figure}
\end{figure*}

Figure \ref{fig:process_figure} shows an overview of the work presented in this paper. We first selected datasets for each task and language pair. With those datasets, we fine-tuned different base models pre-trained for each language, selecting those with best performance. The best models are released as part of the library and can be easily used by researchers and practitioners, through a very simple API.

The rest of the paper is organized as follows. In Section \ref{sec:previous_work} we review the related work. In Section \ref{sec:tasks} we describe the tasks considered and the datasets we used. In Section \ref{sec:method} we describe the pre-trained models we fine-tuned for the task-specific datasets we selected. Section \ref{sec:results} presents the comparison of the performance of the considered models and tasks, including an assessment of fairness. A comparison of \pysentimiento{} with other open-source tools is presented in Section \ref{sec:comparison}.

Finally, in Section \ref{sec:conclusions} we present our conclusions and future work.

\section{Previous Work}

\label{sec:previous_work}
Transfer learning and pre-trained language models became key components in state-of-the-art NLP tasks in the past years \citep{radford2018improving,howard2018universal}. \bert{} \citep{devlin2018bert} is a neural bidirectional language model trained on the Masked-Language-Model (MLM) task and in the Next-Sentence-Prediction (NSP) task. The architecture of \bert{} is based on the Transformer \citep{vaswani2017attention}, a neural network architecture that uses self-attention to compute representations of words in a sentence. \bert{} is pre-trained on a large corpus of text ---such as Wikipedia--- and it is then fine-tuned on a specific downstream task such as Question Answering or Sentiment Analysis. \roberta{} \citep{liu2019roberta} is an optimized pre-training approach. These models supposed great breakthroughs in the performance on several NLP benchmarks \citep{wang-2018-glue, wang2018superglue}. \cite{nozza2020mask} provides a good overview of the \bert{}-based language models.

The advent of language models based on transformers has led to the development of models trained on corpora targeting specific domains of interest rather than generic texts such as Wikipedia or news. SciBERT \citep{beltagy-etal-2019-scibert} is a \bert{} model trained on scientific texts, while MediBERT \citep{rasmy2021med} was trained on medical documents. AlBERTo \citep{polignano2019alberto} was among the first models trained on tweets, specifically in the Italian language. Similarly, \bertweet{} \citep{nguyen2020bertweet} is a \roberta{} model trained on approximately 850 million English tweets, some of which are related to the COVID-19 pandemic. \robertuito{} \citep{perez2022robertuito} is a \roberta{} model trained on 600 million tweets, mostly in Spanish but also featuring some English, Portuguese, and other languages.

Most popular opinion-mining toolkits still rely on rule-based or lexicon-based methodologies. One example of this is \emph{VADER} \citep{hutto2014vader}, a rule-based library specially crafted for Social Media in English, providing multilingual support through translation from the target language. Another notable toolkit is \emph{Textblob} \footnote{\url{https://github.com/sloria/TextBlob/}}, which leverages classic machine learning algorithms, such as Naive Bayes and SVMs, to perform sentiment analysis in English, albeit with some adaptations for other languages. Simultaneously to our work, \textit{TweetNLP} \citep{camacho-2022-tweetnlp} was published, a library for different opinion-mining tasks in Twitter using transformer-based models. However, most of its models are based on multilingual pre-trained models, which have been shown to be suboptimal due to the \emph{the curse of multilinguality} \citep{conneau-2020-unsupervised}.

Huggingface's \emph{transformers} \citep{wolf2019huggingface} has emerged as a widely used open-source tool for natural language processing (NLP) tasks, providing state-of-the-art transformer-based models and tools, making it easier to implement and experiment with NLP models. The library is built on top of PyTorch \cite{paszke2017automatic}, a popular deep learning framework, which allows for efficient and easy implementation of transformer models on both CPUs and GPUs.

\section{Tasks and datasets}
\label{sec:tasks}

\begin{table*}[ht!]
    \centering
    \begin{tabular}{c lll l}
        Language                    & Tasks                 & Type of task & Dataset                        & \#Instances \\
        \hline
        \multirow{4}{*}{Spanish}    & Sentiment analysis    & Multiclass   & TASS 2020 Task A               & $14,509$    \\
                                    & Emotion analysis      & Multiclass   & TASS 2020 Task B               & $8,409$     \\
                                    & Irony detection       & Binary       & IrosVA 2019                    & $9,000$     \\
                                    & Hate speech detection & Multilabel   & HatEval                        & $6,600$     \\
        \hline \rule{0pt}{1.2em}
        \multirow{4}{*}{English}    & Sentiment analysis    & Multiclass   & SemEval 2017 Task 4            & $61,929$    \\
                                    & Emotion analysis      & Multiclass   & GoEmotions                     & $54,263$    \\
                                    & Irony detection       & Binary       & SemEval 2018 Task 3            & $4,601$     \\
                                    & Hate speech detection & Multilabel   & HatEval                        & $13,000$    \\
        \hline \rule{0pt}{1.2em}
        \multirow{4}{*}{Italian}    & Sentiment analysis    & Multilabel   & SENTIPolc 2016                 & $9,410$     \\
                                    & Emotion analysis      & Multiclass   & FEEL-IT                        & $2,037$     \\
                                    & Irony detection       & Binary       & SENTIPolc 2016                 & $9,410$     \\
                                    & Hate speech detection & Multilabel   & HaSpeeDe 2020                  & $9,273$     \\
        \hline \rule{0pt}{1.2em}
        \multirow{4}{*}{Portuguese} & Sentiment analysis    & Multiclass   & Brazilian Portuguese dataset   & $15,000$    \\
                                    & Emotion analysis      & Multiclass   & Translated Go-Emotions         & $53,705$    \\
                                    & Irony detection       & Multi-label  & IDPT 2021                      & $15,212$    \\
                                    & Hate speech detection & Binary       & Hierarchically labeled dataset & $5,668$     \\

        \hline
    \end{tabular}
    \caption{Evaluation tasks for \pysentimiento{}. Tasks are grouped by language, and the type of task is indicated. The dataset used for each task is also indicated.}
    \label{tab:evaluation_settings}
\end{table*}

We tackled four different opinion-mining tasks for \pysentimiento{}: sentiment analysis, emotion detection, hate speech detection, and irony detection. For each of these tasks across the considered languages (Spanish, English, Portuguese and Italian), we selected representative datasets of the task and the language, whenever possible.

\textbf{Sentiment analysis} stands as one of the first and most popular tasks in opinion mining, having attracted a lot of attention since the emergence of the World Wide Web \citep{pang2008opinion}. In its fundamental form, sentiment analysis is represented as polarity detection, where the goal is to predict whether a text has an overall sentiment --- be it positive, negative, or neutral.


\textbf{Emotion detection} is a more recent task that gained popularity in the past years and can be considered as a refinement of polarity detection. An emotion can be described as a state involving physiological, subjective, and expressive components enabling an individual to situate within a social and moral order \citep{keltner2019emotional}. These emotions can be described in terms of six basic building blocks: \emph{anger}, \emph{disgust}, \emph{fear}, \emph{joy}, \emph{sadness}, \emph{surprise} \citep{ekman1999basic}. Detecting emotions in a text can be posed as a classification task consisting of the prediction of the emotion conveyed in a text. In some cases, the labels can be expressed as a combination of the six basic emotions or using a bigger set of fine-grained emotions. This task is inherently more complex than sentiment analysis, as it requires a deeper comprehension of the text --- and it is also more subjective, as different individuals may interpret the emotions quite differently.

\textbf{Hate speech} can be described as speech containing violence towards an individual or a group of individuals, according to certain characteristics protected by international treaties, such as gender, race, language, and others \citep{article192015}. In recent years, hate speech has become problematically relevant due to its intensity and its prevalence on social media, and this phenomenon has been associated with stress and depression of victims \citep{saha2019prevalence}, and also to the settle of a hostile and dehumanizing environment for immigrants, sexual and religious minorities, as well as other vulnerable groups \citep{bilewicz2020hate}.  As a result, there has been a growing interest in detecting and preventing hate speech. There are several approaches to this task, including binary classification, which predicts whether a text contains hate speech or not, and multi-label classification, which predicts whether a text contains a specific type of hate speech (such as racism or sexism) or particular features such as targeting a specific individual or containing calls to action \citep{poletto2021resources}.

\textbf{Irony detection} refers to detecting whether a text is ironic, considering irony as a rhetorical device where the intended meaning of a statement is the opposite of its literal meaning. Detecting whether a text is ironic constitutes a challenging semantic task known as \textbf{irony detection}. Being able to detect irony requires a deep comprehension of the text and is notably subjective, as individuals with varying cultural backgrounds may interpret the irony differently \citep{Frenda2023}. This task is usually framed as a binary classification problem, where the goal is to predict whether a text is ironic or not. In some cases, the task is posed as a multi-label classification problem, where the goal is to predict the type of irony if any \citep{van-hee-etal-2018-semeval}.

We also considered other Social NLP tasks for \pysentimiento{} such as POS tagging, NER, and other opinion-mining tasks such as targeted sentiment analysis, but as they are not the focus of this work, we refer the reader to the Appendix \ref{app:other_tasks} for more details.

\subsection{Datasets}

In this section, we describe the datasets used for each of the aforementioned tasks.
The criteria for selecting these datasets is giving priority to chosing those presented at popular shared tasks in the selected languages, such as \textit{SemEval}, \textit{IberLEF} and \textit{EvalITA}, among others. In most of the cases, however, the chosen datasets were the only dataset available for a given task and language.

We considered polarity detection at its most standard form: given a tweet, predict whether it is positive, negative or neutral. For this task in English, we used the \emph{SemEval 2017} Task 4 Subtask 1 \citep{rosenthal-2017-semeval} dataset, which contains 61,900 tweets annotated for polarity detection. In Spanish, we relied on the TASS 2020 \citep{garcia2020overview} dataset, which is also separated according to different dialects of Spanish. For the purpose of benchmarking our models, we disregarded these distinctions and merged all the data into a single dataset, with respective train, dev, and test splits. For Portuguese, we used a dataset of Brazilian Portuguese tweets, collected from user interactions talking about TV shows \citep{brum2018sentimentportuguese}. Lastly, \textit{SENTIPolc 2016} \citep{barbieri2016overview} was used as the Italian corpus, which contains a dataset for polarity detection with the novelty of adding a fourth class for those tweets with mixed polarities (both positive and negative). This last dataset could be also cast as a multi-label classification task for two variables (positive and negative).

For emotion detection in Spanish, we also leveraged resources from the TASS 2020 workshop, particularly the \emph{EmoEvent} dataset \citep{plaza2020emoevent}. This dataset is labeled with Ekman's six basic emotions (\emph{anger}, \emph{disgust}, \emph{fear}, \emph{joy}, \emph{sadness}, \emph{surprise}) \citep{ekman1999basic} plus a neutral emotion. \emph{EmoEvent} was crafted by retrieving tweets associated with eight different global events from different domains (e.g. political, entertainment, catastrophes or incidents, commemorations, etc.), so emotions are always related to specific events. For Italian, we used the \emph{FEEL-it} dataset \citep{bianchi2021feel}, which is a collection of 2,037 tweets labeled with a subset of Ekman's categories: anger, fear, joy, and sadness. For English, we used the \emph{Go Emotions} dataset \citep{demszky2020goemotions}, which has a fine-grained emotion annotation scheme with 27 emotion categories. Lastly, as Portuguese had no dataset available for emotion detection, we used a translated version of the Go-Emotions dataset\footnote{\url{https://huggingface.co/datasets/antoniomenezes/go_emotions_ptbr}}.

For hate speech detection, we selected the \emph{hatEval} dataset \citep{hateval2019semeval}. Instead of the binary detection task, we analyzed the multi-label setting, which requires predicting three binary, hierarchical variables: \emph{hate speech} (HS), \emph{if hateful, is it against a specific target?} (TR), \emph{if hateful, is it aggressive?} (AG). For Italian, we used the HaSpeeDe 2020 dataset \citep{manuela2020haspeede,bosco2018overview}, which is a collection of tweets labeled for hate speech and stereotype detection. Finally, for Portuguese, we used a dataset of hierarchically-annotated tweets, featuring several characteristics and types of targets \citep{fortuna2019hierarchically}. In all cases, they were considered as multilabel classification tasks.

For irony detection in Spanish, we used IroSVa \citep{ortega2019overview}, a dataset published in 2019, that has the particularity of considering the messages not as isolated texts but with a given context (a headline or a topic). For Italian, we again relied on SENTIPolc, which is also labeled for irony detection. In Portuguese, we used the IDPT 2021 dataset\citep{correa2021overview}. Lastly, we used the SemEval 2018 Task 3 dataset for English \citep{van-hee-etal-2018-semeval}, considering the binary classification task (Task A).

Table \ref{tab:evaluation_settings} summarizes the datasets used for each task and language, as well as the type of classification setting (binary, multiclass, or multilabel) and the number of instances in each dataset. Something important to notice is that the dataset annotations for each pair of language and task are not comparable one-to-one, as some of their characteristics are different. For example, some of them are multi-class tasks while others are multi-label.

\section{Method}
\label{sec:method}

To decide which model to integrate in \pysentimiento{} as the default option to process a given language for a given task, we carried out an extensive evaluation of the performance of each model in the same benchmark, so that meaningful comparisons could be carried out. For each model, we selected the fine-tuned version that obtained the best results in the validation partition of the training corpus during hyperparameter search. Also, we carried out a fairness analysis for the emotion detection task in English, as it is the only task for which we found a suitable corpus with demographic information.

\subsection{Preprocessing}

Preprocessing is crucial when working with Twitter data, which can be quite noisy and in general may contain various non-canonical text elements such as user handles (@username), hashtags, emojis, and misspellings, among others. In their work, \cite{nguyen2020bertweet} attempted two normalization strategies, a soft one that made minor changes such as replacing usernames and hashtags, and a more aggressive one using the ideas of \cite{han2011lexical}. However, the authors found no significant improvement by using the latter normalization strategy.

With this in mind, we adopted an approach similar to the one used both in \cite{nguyen2020bertweet} and in \cite{polignano2019alberto}:

\begin{itemize}
    \item Character repetitions were limited to a max of three
    \item User handles were converted to a special token (\verb|@usuario| in Spanish, \verb|@USER| in English)
    \item Hashtags were replaced by a special token \verb|hashtag| followed by the hashtag text and split into words if possible
    \item Emojis were replaced by their text representation using \emph{emoji} library\footnote{\url{https://github.com/carpedm20/emoji/}}, surrounded by a special token \verb|emoji|.
\end{itemize}

\subsection{Pretrained models}

To select the best-performing models for \pysentimiento{}, we carried out an extensive evaluation focusing on both general-domain models and specialized models tailored for the social-media domain. For English, we tested \bert{} \citep{devlin2018bert}, \roberta{} \citep{liu2019roberta}, and \electra{} \citep{clark2020electra}. Regarding specialized models, we benchmarked \bertweet{} \citep{nguyen2020bertweet}, a pre-trained language model on a large corpus of tweets following the same guidelines as \roberta{}.

In parallel, for the Spanish language, we selected \beto{} \citep{canete2020spanish}, \bertin{} \citep{BERTIN}, \roberta{}$_{ES}$ \citep{gutierrez2022maria}, \electricidad{} and \robertuito{} \citep{perez2022robertuito}. Similarly, for Italian we included BERT and \electra{} for this language \citep{schweter2021bert}, UmBERTo \citep{umberto2019}, AlBERTo \citep{polignano2019alberto}. Finally, for Portuguese, we tested BERTimbau \citep{souza2020bertimbau}, a Portuguese version of BERT, \bertweet{}$_{BR}$ \citep{vianna2023sentiment} and BERTabaPoru \citep{bertabaporu}, these two last corresponding to social-media models. For all these languages, we also tested \robertuito{}, as it showed proficiency in multiple languages closely related to Spanish \citep{perez2022robertuito}. The selected language models were encoder-only because this architecture is best suited to extract relevant features from context for classification tasks. Since these tasks do not require to generate sequence, a decoder is not needed.

Table \ref{tab:pretrained_models} summarizes the models used in the experiments.

\begin{table}[t]
    \centering
    \footnotesize
    \begin{tabular}{c ll}
        Language                    & Pre-trained model & Model family       \\
        \hline
        \multirow{5}{*}{English}    & \bert{}           & ---                \\
                                    & \roberta{}        & ---                \\
                                    & \electra{}        & ---                \\
                                    & \bertweet{}       & Social-media model \\
                                    & \robertuito{}     & Social-media model \\
        \hline \rule{0pt}{1.2em}
        \multirow{5}{*}{English}    & \beto{}           & BERT               \\
                                    & BERTin            & \roberta{}         \\
                                    & \robertaes{}      & \roberta{}         \\
                                    & ELECTRA$_{ES}$    & ELECTRA            \\
                                    & \robertuito{}     & Social-media model \\
        \hline \rule{0pt}{1.2em}
        \multirow{5}{*}{Italian}    & \bert{}$_{IT}$    & BERT based         \\
                                    & UmBERTo           & ELECTRA            \\
                                    & \electra{}$_{IT}$ & \roberta{}         \\
                                    & \alberto{}        & Social-media model \\
                                    & \robertuito{}     & Social-media model \\
        \hline \rule{0pt}{1.2em}
        \multirow{4}{*}{Portuguese} & BERTimbau         & BERT               \\
                                    & BERTweet$_BR$     & Social-media model \\
                                    & BERTabaporu       & Social-media model \\
                                    & \robertuito{}     & Social-media model \\

        \hline
    \end{tabular}
    \caption{Summary of the pre-trained models used in the experiments.}
    \label{tab:pretrained_models}
\end{table}

\subsection{Fine-tuning process}

The classifiers were trained according to the methodology described in \cite{devlin2018bert}, with Adam \citep{kingma2014adam} as the optimizer and a triangular learning rate schedule. We performed an extensive hyperparameter search for each model, task, and language, using the validation set to select the best model. More details about the range of hyperparameter tuning can be found in Appendix \ref{app:training_details}.

To evaluate our models, we ran ten experiments with different random seeds in order to obtain a more robust estimate of the performance. We report the Macro F1 score for each considered task.

\subsection{Fairness evaluation}

AI evaluation based on aggregated metrics provide a general perspective of the models' performance; however, these metrics may hide systematic errors against certain subgroups. Deploying such biased models may result in an amplification of biases against those groups. To prevent this kind of harm, assessing whether a given model is biased is an integral part of the analysis of its performance.

Unfortunately, resources to carry out fairness analysis are even more scarce than those for opinion mining. Such resources need to provide, besides the labels for the targeted task, also some kind of demographic information of the people mentioned, producing or receiving each text. We have been able to find only one such corpus, this is why we have limited our analysis of bias to the emotion detection task in English. Nonetheless, the kind of analysis that we did for English can also be carried out for the rest of languages, translating the dataset, since the properties that this particular dataset relies upon also hold for the rest of languages in this study. More concretely, person names can be consistently associated with gender.

To carry out this analysis, we used the Equity Evaluation Corpus (ECC) dataset  \citep{kiritchenko2018examining}. It comprises $8,640$ English sentences, each one associated with a race, gender, and emotion label. The considered emotions are anger, fear, joy, sadness, and neutral. It is important to notice that this corpus was not gathered ``naturally'' but instead was artificially created, consisting of sentences that mirror each other in structure with minimal variations in names that allow to associate the sentence with different genders or races. For each sentiment, the same number of sentences were associated with each race and gender, i.e., it is a balanced dataset. In spite of the shortcomings of this dataset, being small, artificial, for English only and for a limited number of demographic groups, we believe the fairness assessment provided here constitutes a good guide for each researcher or practitioner to carry out their own assessment.

We used the statistical parity criteria \citep{barocas2019} to assess the fairness of our models. Considering the sensitive attribute A and a classifier $\hat{Y}$, if
$A \in \{0,1\}$ and $\hat{Y} \in \{0,1\}$, we define that the classifier is considered fair under statistical parity definition if
$$P(\hat{Y}=1 \mid A=0)=P(\hat{Y}=1 \mid A=1)$$
i.e., the probability of a positive outcome is the same for the unprivileged and privileged groups. In the literature, a way to quantify the Statistical Parity   is with the Disparate Impact (DI) metric \citep{feldman2015certifying}:
$$DI = \frac{P(\hat{Y}=1 \mid A=0)}{P(\hat{Y}=1 \mid A=1)}$$
When a classifier satisfies the statistical parity fairness definition $DI=1$.

As the emotion detection task considered in English is a multilabel task, we considered each possible combination of model, emotion, and demographic group for the fairness analysis.

\section{Results}
\label{sec:results}

\begin{table*}[t!]
    \centering
    \begin{tabular}{llccccr}
        \hline
        Language           & Model           & Sentiment             & Emotion              & Hate Speech           & Irony                \\
        \hline
        \mr{5}{Spanish}    & BERTin          & $65.3 \pm 0.5$        & $50.2 \pm 2.9$       & $68.7 \pm 1.5$        & $69.3 \pm 1.4      $ \\
                           & BETO            & $67.2 \pm 0.6$        & $52.2 \pm 1.4$       & $73.3 \pm 0.8$        & $71.5 \pm 0.5      $ \\
                           & ELECTRicidad    & $65.3 \pm 0.5$        & $46.3 \pm 2.3$       & $71.8 \pm 1.0$        & $67.1 \pm 2.1      $ \\
                           & RoBERTa$_{es}$  & $67.3 \pm 0.3$        & $53.1 \pm 2.2$       & $73.1 \pm 2.8$        & $71.9 \pm 0.9      $ \\
                           & RoBERTuito      & $\mbf{70.2 \pm 0.2}$  & $\mbf{55.3 \pm 0.8}$ & $\mbf{76.1 \pm 0.5}$  & $\mbf{74.1 \pm 0.7}$ \\
        \hline
        \mr{5}{English}    & BERT            & $69.6 \pm 0.4$        & $42.7 \pm 0.6    $   & $56.0 \pm 0.8$        & $68.1 \pm 2.2      $ \\
                           & ELECTRA         & $70.9 \pm 0.4$        & $37.2 \pm 2.9    $   & $55.6 \pm 0.6$        & $71.3 \pm 1.8      $ \\
                           & RoBERTa         & $70.4 \pm 0.3$        & $\mbf{45.0 \pm 0.9}$ & $55.1 \pm 0.4$        & $70.4 \pm 2.9      $ \\
                           & RoBERTuito      & $69.6 \pm 0.5$        & $43.0 \pm 3.3      $ & $57.5 \pm 0.2$        & $73.9 \pm 1.4      $ \\
                           & BERTweet        & $\mbf{72.0 \pm 0.4}$  & $43.1 \pm 1.8      $ & $\mbf{57.7 \pm 0.7}$  & $\mbf{80.8 \pm 0.7}$ \\

        \hline
        \mr{5}{Italian}    & AlBERTo         & $57.8 \pm 0.7       $ & $72.0 \pm 1.3      $ & $88.1 \pm 0.4       $ & $53.7 \pm 0.6      $ \\
                           & BERT$_{it}$     & $61.4 \pm 0.9       $ & $\mbf{73.6 \pm 4.0}$ & $\mbf{92.4 \pm 0.4} $ & $\mbf{62.0 \pm 4.4}$ \\
                           & ELECTRA$_{it}$  & $\mbf{62.3 \pm 0.7} $ & $64.7 \pm 7.7      $ & $87.8 \pm 3.0       $ & $50.0 \pm 6.5      $ \\
                           & UmBERTo         & $\mbf{62.6 \pm 1.1} $ & $69.7 \pm 4.6      $ & $87.3 \pm 0.4       $ & $60.0 \pm 2.2      $ \\
                           & RoBERTuito      & $55.2 \pm 2.8       $ & $64.1 \pm 3.0      $ & $\mbf{92.6 \pm 0.3} $ & $55.6 \pm 3.6      $ \\
        \hline
        \mr{4}{Portuguese} & BERT$_{pt}$     & $70.0 \pm 0.3      $  & $89.9 \pm 0.2$       & $64.1 \pm 1.1       $ & \mr{4}{---}          \\
                           & BERTweet$_{BR}$ & $\mbf{75.3 \pm 0.5}$  & $\mbf{91.3 \pm 0.4}$ & $55.6 \pm 5.5       $ &                      \\
                           & RoBERTuito      & $71.7 \pm 0.4      $  & $87.6 \pm 0.7$       & $70.0 \pm 2.4       $ &                      \\
                           & BERTabaporu     & $73.8 \pm 0.4      $  & $\mbf{91.6 \pm 0.2}$ & $\mbf{70.3 \pm 3.3} $ &                      \\
        \hline
    \end{tabular}
    \caption{Evaluation results for the classification tasks: hate speech detection, sentiment analysis, emotion analysis and irony detection. Results are expressed as the mean Macro F1 score of 10 runs of the classification experiments, $\pm$ their standard deviation. Bold indicates best performing models}
    \label{tab:evaluation_results}
\end{table*}

The evaluation results for the considered tasks and languages are presented in Table~\ref{tab:evaluation_results}. Compared to other benchmark and the shared tasks overviews \citep[inter alia]{plaza-del-arco2021comparing, camacho-collados2022tweetnlp, perez2022robertuito,garcia2020overview,manuela2020haspeede}, the results are consistent taking into account differences such as the pre-trained models used and the hyperparameter tuning process.

In general, it can be seen that specialized language models for social media show superior performance in most languages: \bertweet{} for English, \bertweet{}${br}$ and BERTabaporu for Portuguese, and \robertuito{} for Spanish. The only exception to this trend is Italian, where the highest-performing model is \bertit{}. The main difference between Twitter-based models in other languages and \alberto{} lies in the pretraining guidelines, with the other languages following the recipe of~\cite{liu2019roberta}. However, an in-depth analysis of the reasons for \alberto{}'s lower performance is beyond the scope of this paper, and should be examined carefully.

Furthermore, \robertuito{} delivers robust performance for most tasks and languages, as highlighted in \cite{perez2022robertuito}. The reason behind this might be the data used on its pretraining, which mainly comprised Spanish tweets along with a substantial amount of English, Portuguese, and other related languages used by Spanish-speaking people on Twitter. This language model could be used as a starting point for other languages, such as Catalan, Galician, or Basque, which are also spoken in Spain (and, with the exception of the latter, are linguistically related to Spanish).

For each task and language, we selected the best performing model to be used in \pysentimiento{}. In the cases where the performance difference was insignificant (e.g. \robertuito{} and \bertit{}), we opted for the monolingual or specialized model. Sadly, we were not able to report results for the Portuguese task of irony detection, as the gold labels for the test set were not available \citep{correa2021overview}.

Table~\ref{tab:fairness_analysis} reports the disparate impact of each model, emotion, and setting (gender or race) for the Emotion detection task in English. We use the four-fifths rule to determine whether the discrimination committed by the models is acceptable \citep{Besse2021}. This rule states that if the
the ratio between the privileged and unprivileged groups is less than 0.8, so the discriminations committed by a model are unacceptable, and the model is considered unfair under the statistical parity criteria.

For all combinations of model and emotion, the DIs are higher than $0.8$, providing no evidence of adverse impact. Most notably, models trained on user-generated content (\bertweet{} and \robertuito{}) show similar behavior to the general-domain models (\bert{}, \roberta{} and \electra{}). This finding suggests that the models trained on social media data do not show a higher bias than the general-domain models. Nonetheless, this should be taken with caution as the ECC corpus might not represent actual deployment context, and also given that large language models have been shown to be highly biased \citep{bender2021dangers,garrido2023maria}.

In spite of these limitations, this analysis serves as a first approach to detect strong biases in opinion mining tools. We recommend using a more representative of the real context of the deployment dataset for a more adequate fairness assessment.

\begin{table*}[t]
    \centering
    \begin{tabular}{ll  ccccc  c}
                       &            & \mc{5}{Emotion} &                                           \\
                       &            & anger           & fear  & joy   & neutral & sadness & mean  \\
                       & Model      &                 &       &       &         &         &       \\
        \hline
        \mr{5}{Race}   & BERT       & 0.952           & 1.119 & 1.055 & 0.845   & 1.033   & 1.001 \\
                       & BERTweet   & 0.995           & 1.130 & 1.022 & 0.879   & 1.041   & 1.013 \\
                       & ELECTRA    & 0.990           & 1.042 & 1.043 & 0.909   & 1.011   & 0.999 \\
                       & RoBERTa    & 0.988           & 1.023 & 0.969 & 0.992   & 1.034   & 1.001 \\
                       & RoBERTuito & 0.966           & 1.016 & 0.992 & 1.000   & 1.014   & 0.998 \\
        \hline
        \mr{5}{Gender} & BERT       & 0.993           & 0.971 & 1.002 & 1.040   & 0.997   & 1.000 \\
                       & BERTweet   & 0.945           & 1.005 & 0.967 & 1.050   & 1.011   & 0.995 \\
                       & ELECTRA    & 0.997           & 0.986 & 0.979 & 1.050   & 0.996   & 1.002 \\
                       & RoBERTa    & 1.013           & 1.059 & 0.988 & 1.001   & 0.962   & 1.005 \\
                       & RoBERTuito & 1.004           & 1.008 & 0.959 & 1.068   & 0.940   & 0.996 \\
        \hline
    \end{tabular}
    \caption{Treatment Equality results obtained for different races (European (E) and Afro-American (AA)) in the Equity Evaluation Corpus (ECC) for different base models for English, fine-tuned for the task of emotion analysis. Treatment Equality provides a compact measure of the distribution of errors, comparing the number of False Negatives and False Positives.}
    \label{tab:fairness_analysis}
\end{table*}

\section{Comparing \pysentimiento{} with other tools}
\label{sec:comparison}

In this section, a comparison of \pysentimiento{} against other available, open-source opinion mining tools is presented. To make this comparison, we selected two of the considered tasks in this work ---sentiment analysis and hate speech detection--- as they are the easiest to compare across the existing tools\footnote{Emotion detection presents the issue regarding a different set of classes for the tools and datasets, which makes it harder to compare. Irony detection is not supported by most of the tools evaluated.}.

To assess the performance of \pysentimiento{}, we chose a group of widely-used open-source tools : \vader{} \citep{hutto2014vader}, \textblob{} \citep{loria2018textblob}, \stanza{} \citep{qi2020stanza}, \tweetnlp{} \citep{camacho-collados2022tweetnlp}, and \flair{} \citep{akbik2019flair}.
To evaluate the performance of the tools, we selected several reference datasets for these tasks. Most of these datasets are in English and, in some cases, we use translated versions of them. While this is clearly suboptimal for Italian, Portuguese and Spanish, it still gives us a rough idea of the performance of the tools for these languages.

For the first considered task, sentiment analysis, we used \textit{Multilingual Amazon Reviews Corpus} \citep{keung2020marc}, the \textit{Stanford Sentiment Treebank} \citep{socher2013recursive} (on its SST-2 or binary version), the \textit{SentEval} dataset \citep{conneau2018senteval}, the \textit{Financial Phrasebank} \citep{malo2014financial}, the \textit{MTEB} task on Twitter Sentiment Analysis \citep{barbieri-2022-xlm}, and the
\textit{Sentiment140} dataset \citep{go2009twitter}. For hate speech detection, we used the \textit{HateCheck} dataset \cite{rottger2021hatecheck} in its original English version and also translated to Spanish, Italian, and Portuguese.

For each task, we selected the most likely class as the prediction (in case the probabilities are provided by the tools). That is, if the classification task is binary, we ignored the neutral probability and only considered the positive and negative probabilities. If probabilities were not provided and the tool returns a non-binary output, we toss a coin to decide the class. In case the case of \flair{}, we could only evaluate it on binary datasets. Macro F1 Score was used as the evaluation metric for both tasks.

Appendix \ref{app:comparison} provides a more in-depth description of the datasets and tools used in this comparison.

\subsection{Comparison results}

\begin{table*}
    \centering
    \begin{tabular}{ll cccccc}
        \hline
                               & Dataset               & \pysentimiento{} & \tweetnlp{}     & \stanza{} & \flair{}        & \textblob{} & \vader{} \\
        \hline
        \multirow[c]{6}{*}{en} & amazon                & $\mathbf{63.2}$  & $60.7$          & $52.9$    & $-   $          & $45.0$      & $19.0$   \\
                               & financial\_phrasebank & $\mathbf{68.2}$  & $58.8$          & $44.1$    & $-   $          & $44.9$      & $25.1$   \\
                               & mteb                  & $\mathbf{75.7}$  & $75.5$          & $51.9$    & $-   $          & $47.4$      & $20.3$   \\
                               & sent\_eval            & $\mathbf{88.3}$  & $81.3$          & $77.4$    & $85.1$          & $67.8$      & $73.0$   \\
                               & sentiment140          & $85.8$           & $\mathbf{86.7}$ & $68.5$    & $-   $          & $63.3$      & $26.0$   \\
                               & sst2                  & $87.9$           & $79.7$          & $85.0$    & $\mathbf{93.6}$ & $65.9$      & $66.8$   \\
        \hline
        \multirow[c]{2}{*}{es} & amazon                & $\mathbf{61.8}$  & $21.5$          & $48.5$    & -               & -           & -        \\
                               & mteb                  & $\mathbf{76.8}$  & $31.6$          & $60.1$    & -               & -           & -        \\
        \hline
        \multirow[c]{2}{*}{it} & feel\_it              & $\mathbf{88.2}$  & $58.7$          & -         & -               & -           & -        \\
                               & mteb                  & $\mathbf{45.6}$  & $34.0$          & -         & -               & -           & -        \\
        \hline
        pt                     & mteb                  & $\mathbf{88.3}$  & $38.2$          & -         & -               & -           & -        \\
        \hline                                                                                                                                     \\
    \end{tabular}

    \caption{Comparison of \pysentimiento{} against other tools for sentiment analysis. Results are expressed as Macro F1 Scores. When a tool does not support a language or it is only available for positive/negative classification, we mark it with a dash.}
    \label{tab:sentiment_comparison}
\end{table*}

Table \ref{tab:sentiment_comparison} shows the results of the comparison for sentiment analysis. \pysentimiento{} outperforms the other tools in most of the datasets when this comparison is possible. Just in the \textit{Sentiment140} dataset, \pysentimiento{} is behind \tweetnlp{}, which in general is the runner-up in performance. The only case in which another tool beats \pysentimiento{} by a significant margin is the \textit{SST-2} dataset, where \flair{} achieves a Macro F1 Score of $93.6$ compared to \pysentimiento{}'s $87.9$. While \flair{} is supposed to be trained on the \textit{IMDB} dataset \citep{maas2011learning}, it might have also been trained on the SST-2 dataset, which would explain the outstanding performance\footnote{Indeed, the \flair{} documentation states that its sentiment analysis model is trained on 'movie and product reviews', something that also matches the SST-2 dataset and is not exactly just the IMDB dataset. See \url{https://flairnlp.github.io/docs/tutorial-basics/tagging-sentiment}.}.

In the case of hate speech detection, Table \ref{tab:hate_comparison} shows the results of the comparison. \pysentimiento{} is ahead of \tweetnlp{} in all the languages evaluated with the exception of English, in which \tweetnlp{} achieves a Macro F1 Score of $68.6$ compared to \pysentimiento{}'s $52.1$. This is something unexpected, as both libraries use HatEval \citep{hateval2019semeval} to train and evaluate their models, and the evaluation results in Section \ref{sec:results} and the ones provided by \tweetnlp{}'s  authors \citep{camacho-2022-tweetnlp} show a similar pattern: \bertweet{} shows a better performance in English than other pre-trained models, including XLM-T, the pretrained model used by \tweetnlp{}. Thus, a possible reason behind this might be the pre-trained language model inducing a low performance in the funcionality tests of \textit{HateCheck}.

To wrap up, \pysentimiento{} achieves better results than its competitors, offering also multilingual support for several opinion mining tasks. Compared to \tweetnlp{} ---the runner-up--- \pysentimiento{} outperforms it in most of the cases, but this might be attributed to \emph{the curse of multilinguality} \citep{conneau-2020-unsupervised}. We can see in this evaluation that the performance of \tweetnlp{} is suboptimal for languages other than English. While \pysentimiento{} offers support for a limited number of languages, the evaluation shows that selecting the right pre-trained model for the task and language leads to better results than using a general, one-for-all, multilingual model.

\begin{table}
    \centering
    \begin{tabular}{l c c}
        \hline
        Language   & \pysentimiento{} & \tweetnlp{}     \\
        \hline
        English    & $52.1$           & $\mathbf{68.6}$ \\
        Spanish    & $\mathbf{66.7}$  & $26.1$          \\
        Italian    & $\mathbf{53.1}$  & $25.4$          \\
        Portuguese & $\mathbf{43.6}$  & $26.2$          \\
        \hline
    \end{tabular}
    \label{tab:hate_comparison}
    \caption{Comparison of \pysentimiento{} against TweetNLP for hate speech detection, using the \textit{HateCheck}. Results are expressed as Macro F1 Scores.}
\end{table}

\section{Conclusions and Future Work}
\label{sec:conclusions}

In this work, we presented \pysentimiento{}, a multilingual toolkit for extracting opinions from social media text. We evaluated the performance of several pre-trained language models for four languages (Spanish, English, Italian, and Portuguese) and four different tasks (sentiment analysis, emotion analysis, hate speech detection, and irony detection). From this evaluation, we selected the best models and put them under an easy-to-use interface in Python for processing and analyzing this kind of text, expecting this will help researchers interested in opinion mining from social networks. Lastly, we compared the performance of \pysentimiento{} with other open-source tools for opinion mining, showing that our toolkit has a state-of-the-art performance in most of the tasks and languages considered.

We also carried out a small fairness assessment. Although this assessment is very small and also limited to English, we provide it as a step-by-step procedure of how practitioners can diagnose relevant biases for their context of application before deployment, thus preventing possible harm to the target population of this tool. Concerning the race analysis, this analysis is not straightforward and extendible to other languages, e.g., in Spanish people's names are not as unequivocally related to races as in English \cite{fort-etal-2024-stereotypical-mileage}. But we consider that it is very valuable to show that this approach does not apply only to gender but also to other social groups. Given a clear delimitation of the privileged/unprivileged social groups and its textual manifestations, for a given specific language and task, it is totally feasible to select a metric that lets you understand if the harmful errors are equally distributed between privileged and unprivileged populations.

Such delimitations and associated linguistic resources are not utopic. Indeed, they are growing (e.g., the multilingual crows pairs initiative \cite{fort-etal-2024-stereotypical-mileage}), alas not yet providing specific annotations for sentiment, to our knowledge. We expect that such resources will be shortly available for sentiment as well, and in that moment the methodologic illustration provided in this paper will be very valuable for others to carry out this research. If we are still able to carry out this research (the situation of scientific institutions in Argentina permitting), we will be most than happy to carry out the corresponding experiments and share results with the community.

We make our code and models publicly available at \textit{GitHub} \footnote{\url{https://github.com/pysentimiento/pysentimiento/}} and the \textit{huggingface} hub \footnote{\url{https://huggingface.co/pysentimiento}}. We plan to extend \pysentimiento{} to other languages and tasks, and also provide more information extraction utilities, as \pysentimiento{} only supports NER for Spanish and English. Another line of work should consider --and analyze-- analyzing contextualized information and not just isolated sentences.


\section*{Acknowledgements}

We want to thank the CCAD – Universidad Nacional de Córdoba\footnote{\url{https://ccad.unc.edu.ar/}}, part of SNCAD – MinCyT, República Argentina, for providing access to the computational resources used in this work.

\bibliography{biblio}

\begin{thebibliography}{82}
\providecommand{\natexlab}[1]{#1}
\providecommand{\url}[1]{{#1}}
\providecommand{\urlprefix}{URL }
\providecommand{\doi}[1]{\url{https://doi.org/#1}}
\providecommand{\eprint}[2][]{\url{#2}}
 \bibcommenthead

\bibitem[{Aguilar et~al(2020)Aguilar, Kar, and Solorio}]{aguilar2020lince}
Aguilar G, Kar S, Solorio T (2020) Lince: A centralized benchmark for
  linguistic code-switching evaluation. In: Proceedings of the 12th Language
  Resources and Evaluation Conference, pp 1803--1813

\bibitem[{Akbik et~al(2019)Akbik, Bergmann, Blythe, Rasul, Schweter, and
  Vollgraf}]{akbik2019flair}
Akbik A, Bergmann T, Blythe D, et~al (2019) {{FLAIR}}: {{An Easy-to-Use
  Framework}} for {{State-of-the-Art NLP}}. In: Ammar W, Louis A, Mostafazadeh
  N (eds) Proceedings of the 2019 {{Conference}} of the {{North American
  Chapter}} of the {{Association}} for {{Computational Linguistics}}
  ({{Demonstrations}}). Association for Computational Linguistics, Minneapolis,
  Minnesota, pp 54--59, \doi{10.18653/v1/N19-4010}

\bibitem[{Plaza~del Arco et~al(2020)Plaza~del Arco, Strapparava, Urena~Lopez,
  and Martin}]{plaza2020emoevent}
Plaza~del Arco FM, Strapparava C, Urena~Lopez LA, et~al (2020) {E}mo{E}vent: A
  multilingual emotion corpus based on different events. In: Proceedings of The
  12th Language Resources and Evaluation Conference. European Language
  Resources Association, Marseille, France, pp 1492--1498,
  \urlprefix\url{https://www.aclweb.org/anthology/2020.lrec-1.186}

\bibitem[{{Article 19}(2015)}]{article192015}
{Article 19} (2015) Hate speech explained: A toolkit. Tech. rep., Article 19,
  London, UK, London, UK

\bibitem[{Barbieri et~al(2016)Barbieri, Basile, Croce, Nissim, Novielli, and
  Patti}]{barbieri2016overview}
Barbieri F, Basile V, Croce D, et~al (2016) Overview of the evalita 2016
  sentiment polarity classification task. In: Proceedings of third Italian
  conference on computational linguistics (CLiC-it 2016) \& fifth evaluation
  campaign of natural language processing and speech tools for Italian. Final
  Workshop (EVALITA 2016)

\bibitem[{Barbieri et~al(2022)Barbieri, Espinosa~Anke, and
  Camacho-Collados}]{barbieri-2022-xlm}
Barbieri F, Espinosa~Anke L, Camacho-Collados J (2022) {XLM}-{T}: Multilingual
  language models in {T}witter for sentiment analysis and beyond. In:
  Proceedings of the Thirteenth Language Resources and Evaluation Conference.
  European Language Resources Association, Marseille, France, pp 258--266,
  \urlprefix\url{https://aclanthology.org/2022.lrec-1.27}

\bibitem[{Barocas et~al(2019)Barocas, Hardt, and Narayanan}]{barocas2019}
Barocas S, Hardt M, Narayanan A (2019) Fairness and Machine Learning:
  Limitations and Opportunities. fairmlbook.org,
  \url{http://www.fairmlbook.org}

\bibitem[{Basile et~al(2019)Basile, Bosco, Fersini, Nozza, Patti, Rangel,
  Rosso, and Sanguinetti}]{hateval2019semeval}
Basile V, Bosco C, Fersini E, et~al (2019) Semeval-2019 task 5: Multilingual
  detection of hate speech against immigrants and women in twitter. In:
  Proceedings of the 13th International Workshop on Semantic Evaluation
  (SemEval-2019). Association for Computational Linguistics

\bibitem[{Beltagy et~al(2019)Beltagy, Lo, and
  Cohan}]{beltagy-etal-2019-scibert}
Beltagy I, Lo K, Cohan A (2019) {S}ci{BERT}: A pretrained language model for
  scientific text. In: Proceedings of the 2019 Conference on Empirical Methods
  in Natural Language Processing and the 9th International Joint Conference on
  Natural Language Processing (EMNLP-IJCNLP). Association for Computational
  Linguistics, Hong Kong, China, pp 3615--3620, \doi{10.18653/v1/D19-1371},
  \urlprefix\url{https://aclanthology.org/D19-1371}

\bibitem[{Bender et~al(2021)Bender, Gebru, McMillan-Major, and
  Shmitchell}]{bender2021dangers}
Bender EM, Gebru T, McMillan-Major A, et~al (2021) On the dangers of stochastic
  parrots: Can language models be too big? In: Proceedings of the 2021 ACM
  conference on fairness, accountability, and transparency, pp 610--623

\bibitem[{Besse et~al(2022)Besse, del Barrio, Gordaliza, Loubes, and
  Risser}]{Besse2021}
Besse P, del Barrio E, Gordaliza P, et~al (2022) A survey of bias in machine
  learning through the prism of statistical parity. The American Statistician
  76(2):188--198. \doi{10.1080/00031305.2021.1952897},
  \urlprefix\url{https://doi.org/10.1080/00031305.2021.1952897}

\bibitem[{Bianchi et~al(2021)Bianchi, Nozza, and Hovy}]{bianchi2021feel}
Bianchi F, Nozza D, Hovy D (2021) {"FEEL-IT: Emotion and Sentiment
  Classification for the Italian Language"}. In: Proceedings of the 11th
  Workshop on Computational Approaches to Subjectivity, Sentiment and Social
  Media Analysis. Association for Computational Linguistics

\bibitem[{Biewald(2020)}]{wandb}
Biewald L (2020) Experiment tracking with weights and biases.
  \urlprefix\url{https://www.wandb.com/}, software available from wandb.com

\bibitem[{Bilewicz and Soral(2020)}]{bilewicz2020hate}
Bilewicz M, Soral W (2020) Hate speech epidemic. the dynamic effects of
  derogatory language on intergroup relations and political radicalization.
  Political Psychology 41:3--33

\bibitem[{Bosco et~al(2018)Bosco, Felice, Poletto, Sanguinetti, Maurizio
  et~al}]{bosco2018overview}
Bosco C, Felice D, Poletto F, et~al (2018) Overview of the evalita 2018 hate
  speech detection task. In: Ceur workshop proceedings, CEUR, pp 1--9

\bibitem[{Brum and das Gra\c{c}as
  Volpe~Nunes(2018)}]{brum2018sentimentportuguese}
Brum H, das Gra\c{c}as Volpe~Nunes M (2018) {Building a Sentiment Corpus of
  Tweets in Brazilian Portuguese}. In: chair) NCC, Choukri K, Cieri C, et~al
  (eds) Proceedings of the Eleventh International Conference on Language
  Resources and Evaluation (LREC 2018). European Language Resources Association
  (ELRA), Miyazaki, Japan

\bibitem[{Camacho-collados et~al(2022)Camacho-collados, Rezaee, Riahi, Ushio,
  Loureiro, Antypas, Boisson, Espinosa~Anke, Liu, and
  Mart{\'\i}nez~C{\'a}mara}]{camacho-2022-tweetnlp}
Camacho-collados J, Rezaee K, Riahi T, et~al (2022) {T}weet{NLP}: Cutting-edge
  natural language processing for social media. In: Proceedings of the 2022
  Conference on Empirical Methods in Natural Language Processing: System
  Demonstrations. Association for Computational Linguistics, Abu Dhabi, UAE, pp
  38--49, \urlprefix\url{https://aclanthology.org/2022.emnlp-demos.5}

\bibitem[{{Camacho-collados} et~al(2022){Camacho-collados}, Rezaee, Riahi,
  Ushio, Loureiro, Antypas, Boisson, Espinosa~Anke, Liu, and
  Mart{\'i}nez~C{\'a}mara}]{camacho-collados2022tweetnlp}
{Camacho-collados} J, Rezaee K, Riahi T, et~al (2022) {{TweetNLP}}:
  {{Cutting-Edge Natural Language Processing}} for {{Social Media}}. In: Che W,
  Shutova E (eds) Proceedings of the 2022 {{Conference}} on {{Empirical
  Methods}} in {{Natural Language Processing}}: {{System Demonstrations}}.
  Association for Computational Linguistics, Abu Dhabi, UAE, pp 38--49,
  \doi{10.18653/v1/2022.emnlp-demos.5}

\bibitem[{Canete et~al(2020)Canete, Chaperon, Fuentes, Ho, Kang, and
  P{\'e}rez}]{canete2020spanish}
Canete J, Chaperon G, Fuentes R, et~al (2020) Spanish pre-trained bert model
  and evaluation data. Pml4dc at iclr 2020:2020

\bibitem[{Clark et~al(2020)Clark, Luong, Le, and Manning}]{clark2020electra}
Clark K, Luong MT, Le QV, et~al (2020) Electra: Pre-training text encoders as
  discriminators rather than generators. \eprint{2003.10555}

\bibitem[{Conneau and Kiela(2018)}]{conneau2018senteval}
Conneau A, Kiela D (2018) {{SentEval}}: {{An Evaluation Toolkit}} for
  {{Universal Sentence Representations}}. In: Calzolari N, Choukri K, Cieri C,
  et~al (eds) Proceedings of the {{Eleventh International Conference}} on
  {{Language Resources}} and {{Evaluation}} ({{LREC}} 2018). European Language
  Resources Association (ELRA), Miyazaki, Japan

\bibitem[{Conneau et~al(2020)Conneau, Khandelwal, Goyal, Chaudhary, Wenzek,
  Guzm{\'a}n, Grave, Ott, Zettlemoyer, and
  Stoyanov}]{conneau-2020-unsupervised}
Conneau A, Khandelwal K, Goyal N, et~al (2020) Unsupervised cross-lingual
  representation learning at scale. In: Proceedings of the 58th Annual Meeting
  of the Association for Computational Linguistics. Association for
  Computational Linguistics, Online, pp 8440--8451,
  \doi{10.18653/v1/2020.acl-main.747},
  \urlprefix\url{https://aclanthology.org/2020.acl-main.747}

\bibitem[{Corr{\^e}a et~al(2021)Corr{\^e}a, Coelho, Santos, and
  de~Freitas}]{correa2021overview}
Corr{\^e}a UB, Coelho L, Santos L, et~al (2021) Overview of the idpt task on
  irony detection in portuguese at iberlef 2021. Procesamiento del Lenguaje
  Natural 67:269--276

\bibitem[{da~Costa et~al(2023)da~Costa, Pavan, dos Santos, da~Silva, and
  Paraboni}]{bertabaporu}
da~Costa PB, Pavan MC, dos Santos WR, et~al (2023) {BERTabaporu: assessing a
  genre-specific language model for Portuguese NLP}. In: Recents Advances in
  Natural Language Processing ({RANLP-2023}), Varna, Bulgaria

\bibitem[{Demszky et~al(2020)Demszky, Movshovitz-Attias, Ko, Cowen, Nemade, and
  Ravi}]{demszky2020goemotions}
Demszky D, Movshovitz-Attias D, Ko J, et~al (2020) {GoEmotions: A Dataset of
  Fine-Grained Emotions}. In: 58th Annual Meeting of the Association for
  Computational Linguistics (ACL)

\bibitem[{Devlin et~al(2019)Devlin, Chang, Lee, and Toutanova}]{devlin2018bert}
Devlin J, Chang MW, Lee K, et~al (2019) {BERT}: Pre-training of deep
  bidirectional transformers for language understanding. In: Proceedings of the
  2019 Conference of the North {A}merican Chapter of the Association for
  Computational Linguistics: Human Language Technologies, Volume 1 (Long and
  Short Papers). Association for Computational Linguistics, Minneapolis,
  Minnesota, pp 4171--4186, \doi{10.18653/v1/N19-1423},
  \urlprefix\url{https://aclanthology.org/N19-1423}

\bibitem[{Ekman(1999)}]{ekman1999basic}
Ekman P (1999) Basic emotions. Handbook of cognition and emotion 98(45-60):16

\bibitem[{Feldman et~al(2015)Feldman, Friedler, Moeller, Scheidegger, and
  Venkatasubramanian}]{feldman2015certifying}
Feldman M, Friedler SA, Moeller J, et~al (2015) Certifying and removing
  disparate impact. In: proceedings of the 21th ACM SIGKDD international
  conference on knowledge discovery and data mining, pp 259--268

\bibitem[{Fort et~al(2024)Fort, Alonso~Alemany, Benotti, Bezan{\c{c}}on, Borg,
  Borg, Chen, Ducel, Dupont, Ivetta, Li, Mieskes, Naguib, Qian, Radaelli,
  Schmeisser-Nieto, Raimundo~Schulz, Saci, Saidi, Torroba~Marchante, Xie,
  Zanotto, and N{\'e}v{\'e}ol}]{fort-etal-2024-stereotypical-mileage}
Fort K, Alonso~Alemany L, Benotti L, et~al (2024) Your stereotypical mileage
  may vary: Practical challenges of evaluating biases in multiple languages and
  cultural contexts. In: Calzolari N, Kan MY, Hoste V, et~al (eds) Proceedings
  of the 2024 Joint International Conference on Computational Linguistics,
  Language Resources and Evaluation (LREC-COLING 2024). ELRA and ICCL, Torino,
  Italia, pp 17,764--17,769,
  \urlprefix\url{https://aclanthology.org/2024.lrec-main.1545}

\bibitem[{Fortuna et~al(2019)Fortuna, da~Silva, Wanner, Nunes
  et~al}]{fortuna2019hierarchically}
Fortuna P, da~Silva JR, Wanner L, et~al (2019) A hierarchically-labeled
  portuguese hate speech dataset. In: Proceedings of the third workshop on
  abusive language online, pp 94--104

\bibitem[{Frenda et~al(2023)Frenda, Lo, Casola, Scarlini, Marco, Basile, and
  Bernardi}]{Frenda2023}
Frenda S, Lo SM, Casola S, et~al (2023) Does anyone see the irony here?
  analysis of perspective-aware model predictions in irony detection. In: ECAI
  2023 Workshop on Perspectivist Approaches to NLP

\bibitem[{García-Vega et~al(2020)García-Vega, Díaz-Galiano,
  García-Cumbreras, Plaza-Del-Arco, Montejo-Ráez, Zafra, Martínez-Cámara,
  Aguilar, Antonio, Cabezudo, Chiruzzo, and Moctezuma}]{garcia2020overview}
García-Vega M, Díaz-Galiano M, García-Cumbreras M, et~al (2020) Overview of
  tass 2020: Introducing emotion detection. Procesamiento del Lenguaje Natural

\bibitem[{Garrido-Mu{\~n}oz et~al(2023)Garrido-Mu{\~n}oz,
  Mart{\'\i}nez-Santiago, and Montejo-R{\'a}ez}]{garrido2023maria}
Garrido-Mu{\~n}oz I, Mart{\'\i}nez-Santiago F, Montejo-R{\'a}ez A (2023) Maria
  and beto are sexist: evaluating gender bias in large language models for
  spanish. Language Resources and Evaluation pp 1--31

\bibitem[{Go et~al(2009)Go, Bhayani, and Huang}]{go2009twitter}
Go A, Bhayani R, Huang L (2009) Twitter sentiment classification using distant
  supervision. CS224N project report, Stanford 1(12):2009

\bibitem[{Godbole et~al(2023)Godbole, Dahl, Gilmer, Shallue, and
  Nado}]{tuningplaybookgithub}
Godbole V, Dahl GE, Gilmer J, et~al (2023) Deep learning tuning playbook.
  \urlprefix\url{https://github.com/google-research/tuning_playbook}, version 1

\bibitem[{Guti{\'e}rrez~Fandi{\~n}o et~al(2022)Guti{\'e}rrez~Fandi{\~n}o,
  Armengol~Estap{\'e}, P{\`a}mies, Llop~Palao, Silveira~Ocampo, Pio~Carrino,
  Armentano~Oller, Rodriguez~Penagos, Gonzalez~Agirre, and
  Villegas}]{gutierrez2022maria}
Guti{\'e}rrez~Fandi{\~n}o A, Armengol~Estap{\'e} J, P{\`a}mies M, et~al (2022)
  Maria: Spanish language models. Procesamiento del Lenguaje Natural 68

\bibitem[{Han and Baldwin(2011)}]{han2011lexical}
Han B, Baldwin T (2011) Lexical normalisation of short text messages: Makn sens
  a {\#}twitter. In: Proceedings of the 49th Annual Meeting of the Association
  for Computational Linguistics: Human Language Technologies. Association for
  Computational Linguistics, Portland, Oregon, USA, pp 368--378,
  \urlprefix\url{https://aclanthology.org/P11-1038}

\bibitem[{Honnibal et~al(2020)Honnibal, Montani, Van~Landeghem, and
  Boyd}]{honnibal2020spacy}
Honnibal M, Montani I, Van~Landeghem S, et~al (2020) {spaCy:
  Industrial-strength Natural Language Processing in Python}.
  \doi{10.5281/zenodo.1212303}

\bibitem[{Horrigan(2008)}]{horrigan2008online}
Horrigan J (2008) Online shopping, pew internet \& american life project
  report. Washington, DC: Pew Research Center pp 1--42

\bibitem[{Howard and Ruder(2018)}]{howard2018universal}
Howard J, Ruder S (2018) Universal language model fine-tuning for text
  classification. Proceedings of the 56th Annual Meeting of the Association for
  Computational Linguistics (Volume 1: Long Papers) pp 328--339.
  \doi{10.18653/v1/P18-1031}, \urlprefix\url{https://aclanthology.org/P18-1031}

\bibitem[{Hu and Liu(2004)}]{hu2004mining}
Hu M, Liu B (2004) Mining and summarizing customer reviews. In: Proceedings of
  the Tenth {{ACM SIGKDD}} International Conference on {{Knowledge}} Discovery
  and Data Mining. Association for Computing Machinery, New York, NY, USA,
  {{KDD}} '04, pp 168--177, \doi{10.1145/1014052.1014073}

\bibitem[{Hutto and Gilbert(2014)}]{hutto2014vader}
Hutto C, Gilbert E (2014) Vader: A parsimonious rule-based model for sentiment
  analysis of social media text. In: Proceedings of the International AAAI
  Conference on Web and Social Media

\bibitem[{Kaur et~al(2020)Kaur, Kaul, and Zadeh}]{kaur2020monitoring}
Kaur S, Kaul P, Zadeh PM (2020) Monitoring the dynamics of emotions during
  covid-19 using twitter data. Procedia Computer Science 177:423--430

\bibitem[{Keltner et~al(2019)Keltner, Sauter, Tracy, and
  Cowen}]{keltner2019emotional}
Keltner D, Sauter D, Tracy J, et~al (2019) Emotional expression: Advances in
  basic emotion theory. Journal of nonverbal behavior 43:133--160

\bibitem[{Keung et~al(2020)Keung, Lu, Szarvas, and Smith}]{keung2020marc}
Keung P, Lu Y, Szarvas G, et~al (2020) The multilingual amazon reviews corpus.
  In: Proceedings of the 2020 Conference on Empirical Methods in Natural
  Language Processing

\bibitem[{Kingma and Ba(2014)}]{kingma2014adam}
Kingma DP, Ba J (2014) Adam: A method for stochastic optimization. arXiv
  preprint arXiv:14126980

\bibitem[{Kiritchenko and Mohammad(2018)}]{kiritchenko2018examining}
Kiritchenko S, Mohammad SM (2018) Examining gender and race bias in two hundred
  sentiment analysis systems. NAACL HLT 2018 p~43

\bibitem[{Liu et~al(2019)Liu, Ott, Goyal, Du, Joshi, Chen, Levy, Lewis,
  Zettlemoyer, and Stoyanov}]{liu2019roberta}
Liu Y, Ott M, Goyal N, et~al (2019) Roberta: A robustly optimized bert
  pretraining approach. arXiv preprint arXiv:190711692

\bibitem[{Loria(2018)}]{loria2018textblob}
Loria S (2018) textblob documentation. Release 015 2

\bibitem[{Maas et~al(2011)Maas, Daly, Pham, Huang, Ng, and
  Potts}]{maas2011learning}
Maas AL, Daly RE, Pham PT, et~al (2011) Learning {{Word Vectors}} for
  {{Sentiment Analysis}}. In: Lin D, Matsumoto Y, Mihalcea R (eds) Proceedings
  of the 49th {{Annual Meeting}} of the {{Association}} for {{Computational
  Linguistics}}: {{Human Language Technologies}}. Association for Computational
  Linguistics, Portland, Oregon, USA, pp 142--150

\bibitem[{Malo et~al(2014)Malo, Sinha, Korhonen, Wallenius, and
  Takala}]{malo2014financial}
Malo P, Sinha A, Korhonen P, et~al (2014) Good debt or bad debt: Detecting
  semantic orientations in economic texts. Journal of the Association for
  Information Science and Technology 65

\bibitem[{Manuela et~al(2020)Manuela, Gloria, Di~Nuovo, Frenda, Stranisci,
  Bosco, Tommaso, Patti, Irene et~al}]{manuela2020haspeede}
Manuela S, Gloria C, Di~Nuovo E, et~al (2020) Haspeede 2@ evalita2020: Overview
  of the evalita 2020 hate speech detection task. In: Proceedings of the
  Seventh Evaluation Campaign of Natural Language Processing and Speech Tools
  for Italian. Final Workshop (EVALITA 2020), CEUR, pp 1--9

\bibitem[{Muennighoff et~al(2023)Muennighoff, Tazi, Magne, and
  Reimers}]{muennighoff2023mteb}
Muennighoff N, Tazi N, Magne L, et~al (2023) {{MTEB}}: {{Massive Text Embedding
  Benchmark}}. In: Vlachos A, Augenstein I (eds) Proceedings of the 17th
  {{Conference}} of the {{European Chapter}} of the {{Association}} for
  {{Computational Linguistics}}. Association for Computational Linguistics,
  Dubrovnik, Croatia, pp 2014--2037, \doi{10.18653/v1/2023.eacl-main.148}

\bibitem[{Nguyen et~al(2020)Nguyen, Vu, and Nguyen}]{nguyen2020bertweet}
Nguyen DQ, Vu T, Nguyen AT (2020) Bertweet: A pre-trained language model for
  english tweets. In: Proceedings of the 2020 Conference on Empirical Methods
  in Natural Language Processing: System Demonstrations, pp 9--14

\bibitem[{Nozza et~al(2020)Nozza, Bianchi, and Hovy}]{nozza2020mask}
Nozza D, Bianchi F, Hovy D (2020) What the [mask]? making sense of
  language-specific bert models. arXiv preprint arXiv:200302912

\bibitem[{Ortega-Bueno et~al(2019)Ortega-Bueno, Rangel,
  Hern{\'a}ndez~Far{\i}as, Rosso, Montes-y G{\'o}mez, and
  Medina~Pagola}]{ortega2019overview}
Ortega-Bueno R, Rangel F, Hern{\'a}ndez~Far{\i}as D, et~al (2019) Overview of
  the task on irony detection in spanish variants. In: Proceedings of the
  Iberian languages evaluation forum (IberLEF 2019), co-located with 34th
  conference of the Spanish Society for natural language processing (SEPLN
  2019). CEUR-WS. org, pp 229--256

\bibitem[{Pang and Lee(2008)}]{pang2008opinion}
Pang B, Lee L (2008) Opinion mining and sentiment analysis. Found Trends Inf
  Retr 2(1–2):1–135. \doi{10.1561/1500000011},
  \urlprefix\url{https://doi.org/10.1561/1500000011}

\bibitem[{Parisi(2019)}]{umberto2019}
Parisi L (2019) Umberto: Italian roberta.
  \url{https://huggingface.co/Musixmatch/umberto-wikipedia-uncased-v1}

\bibitem[{Paszke et~al(2017)Paszke, Gross, Chintala, Chanan, Yang, DeVito, Lin,
  Desmaison, Antiga, and Lerer}]{paszke2017automatic}
Paszke A, Gross S, Chintala S, et~al (2017) Automatic differentiation in
  pytorch. In: NIPS-W

\bibitem[{P{\'e}rez et~al(2022)P{\'e}rez, Furman, Alemany, and
  Luque}]{perez2022robertuito}
P{\'e}rez JM, Furman DA, Alemany LA, et~al (2022) Robertuito: a pre-trained
  language model for social media text in spanish. In: Proceedings of the
  Thirteenth Language Resources and Evaluation Conference, pp 7235--7243

\bibitem[{{Plaza-del-Arco} et~al(2021){Plaza-del-Arco}, {Molina-Gonz{\'a}lez},
  {Ure{\~n}a-L{\'o}pez}, and
  {Mart{\'i}n-Valdivia}}]{plaza-del-arco2021comparing}
{Plaza-del-Arco} FM, {Molina-Gonz{\'a}lez} MD, {Ure{\~n}a-L{\'o}pez} LA, et~al
  (2021) Comparing pre-trained language models for {{Spanish}} hate speech
  detection. Expert Systems with Applications 166:114,120.
  \doi{10.1016/j.eswa.2020.114120}

\bibitem[{Poletto et~al(2021)Poletto, Basile, Sanguinetti, Bosco, and
  Patti}]{poletto2021resources}
Poletto F, Basile V, Sanguinetti M, et~al (2021) Resources and benchmark
  corpora for hate speech detection: a systematic review. Language Resources
  and Evaluation 55:477--523

\bibitem[{Polignano et~al(2019)Polignano, Basile, de~Gemmis, Semeraro, and
  Basile}]{polignano2019alberto}
Polignano M, Basile P, de~Gemmis M, et~al (2019) {AlBERTo: Italian BERT
  Language Understanding Model for NLP Challenging Tasks Based on Tweets}. In:
  Proceedings of the Sixth Italian Conference on Computational Linguistics
  (CLiC-it 2019), vol 2481. CEUR,
  \urlprefix\url{https://www.scopus.com/inward/record.uri?eid=2-s2.0-85074851349&partnerID=40&md5=7abed946e06f76b3825ae5e294ffac14}

\bibitem[{Pérez et~al(2023{\natexlab{a}})Pérez, Luque, Zayat, Kondratzky,
  Moro, Serrati, Zajac, Miguel, Debandi, Gravano, and
  Cotik}]{perez2022contextual}
Pérez JM, Luque FM, Zayat D, et~al (2023{\natexlab{a}}) Assessing the impact
  of contextual information in hate speech detection. IEEE Access
  11:30,575--30,590. \doi{10.1109/ACCESS.2023.3258973}

\bibitem[{Pérez et~al(2023{\natexlab{b}})Pérez, Recart~Zapata,
  Alves~Salgueiro, Furman, and Fernández~Larrosa}]{perez2023targeted}
Pérez JM, Recart~Zapata E, Alves~Salgueiro T, et~al (2023{\natexlab{b}}) A
  spanish dataset for targeted sentiment analysis of political headlines.
  Electronic Journal of SADIO (EJS) 22(1):53--66.
  \urlprefix\url{https://publicaciones.sadio.org.ar/index.php/EJS/article/view/467}

\bibitem[{Qi et~al(2020)Qi, Zhang, Zhang, Bolton, and Manning}]{qi2020stanza}
Qi P, Zhang Y, Zhang Y, et~al (2020) Stanza: {{A Python Natural Language
  Processing Toolkit}} for {{Many Human Languages}}. In: Celikyilmaz A, Wen TH
  (eds) Proceedings of the 58th {{Annual Meeting}} of the {{Association}} for
  {{Computational Linguistics}}: {{System Demonstrations}}. Association for
  Computational Linguistics, Online, pp 101--108,
  \doi{10.18653/v1/2020.acl-demos.14}

\bibitem[{Radford et~al(2018)Radford, Narasimhan, Salimans, and
  Sutskever}]{radford2018improving}
Radford A, Narasimhan K, Salimans T, et~al (2018) Improving language
  understanding by generative pre-training

\bibitem[{Rasmy et~al(2021)Rasmy, Xiang, Xie, Tao, and Zhi}]{rasmy2021med}
Rasmy L, Xiang Y, Xie Z, et~al (2021) Med-bert: pretrained contextualized
  embeddings on large-scale structured electronic health records for disease
  prediction. NPJ digital medicine 4(1):1--13

\bibitem[{Rosa et~al(2022)Rosa, Ponferrada, Romero, Villegas, de~Prado~Salas,
  and Grandury}]{BERTIN}
Rosa JDL, Ponferrada EG, Romero M, et~al (2022) Bertin: Efficient pre-training
  of a spanish language model using perplexity sampling. pp 13--23,
  \urlprefix\url{http://journal.sepln.org/sepln/ojs/ojs/index.php/pln/article/view/6403}

\bibitem[{Rosenthal et~al(2017)Rosenthal, Farra, and
  Nakov}]{rosenthal-2017-semeval}
Rosenthal S, Farra N, Nakov P (2017) {S}em{E}val-2017 task 4: Sentiment
  analysis in {T}witter. In: Proceedings of the 11th International Workshop on
  Semantic Evaluation ({S}em{E}val-2017). Association for Computational
  Linguistics, Vancouver, Canada, pp 502--518, \doi{10.18653/v1/S17-2088},
  \urlprefix\url{https://aclanthology.org/S17-2088}

\bibitem[{R{\"o}ttger et~al(2021)R{\"o}ttger, Vidgen, Nguyen, Waseem, Margetts,
  and Pierrehumbert}]{rottger2021hatecheck}
R{\"o}ttger P, Vidgen B, Nguyen D, et~al (2021) {{HateCheck}}: {{Functional
  Tests}} for {{Hate Speech Detection Models}}. In: Zong C, Xia F, Li W, et~al
  (eds) Proceedings of the 59th {{Annual Meeting}} of the {{Association}} for
  {{Computational Linguistics}} and the 11th {{International Joint Conference}}
  on {{Natural Language Processing}} ({{Volume}} 1: {{Long Papers}}).
  Association for Computational Linguistics, Online, pp 41--58,
  \doi{10.18653/v1/2021.acl-long.4}

\bibitem[{Saha et~al(2019)Saha, Chandrasekharan, and
  De~Choudhury}]{saha2019prevalence}
Saha K, Chandrasekharan E, De~Choudhury M (2019) Prevalence and psychological
  effects of hateful speech in online college communities. In: Proceedings of
  the 10th ACM conference on web science, pp 255--264

\bibitem[{Sanh et~al(2020)Sanh, Debut, Chaumond, and Wolf}]{sanh2020distilbert}
Sanh V, Debut L, Chaumond J, et~al (2020) {{DistilBERT}}, a distilled version
  of {{BERT}}: Smaller, faster, cheaper and lighter.
  \doi{10.48550/arXiv.1910.01108}, \eprint{1910.01108}

\bibitem[{Schweter(2021)}]{schweter2021bert}
Schweter S (2021) dbmdz bert models. \url{https://github.com/dbmdz/berts}

\bibitem[{Socher et~al(2013)Socher, Perelygin, Wu, Chuang, Manning, Ng, and
  Potts}]{socher2013recursive}
Socher R, Perelygin A, Wu J, et~al (2013) Recursive {{Deep Models}} for
  {{Semantic Compositionality Over}} a {{Sentiment Treebank}}. In: Yarowsky D,
  Baldwin T, Korhonen A, et~al (eds) Proceedings of the 2013 {{Conference}} on
  {{Empirical Methods}} in {{Natural Language Processing}}. Association for
  Computational Linguistics, Seattle, Washington, USA, pp 1631--1642

\bibitem[{Souza et~al(2020)Souza, Nogueira, and Lotufo}]{souza2020bertimbau}
Souza F, Nogueira R, Lotufo R (2020) {BERT}imbau: pretrained {BERT} models for
  {B}razilian {P}ortuguese. In: 9th Brazilian Conference on Intelligent
  Systems, {BRACIS}, Rio Grande do Sul, Brazil, October 20-23 (to appear)

\bibitem[{Van~Hee et~al(2018)Van~Hee, Lefever, and
  Hoste}]{van-hee-etal-2018-semeval}
Van~Hee C, Lefever E, Hoste V (2018) {S}em{E}val-2018 task 3: Irony detection
  in {E}nglish tweets. In: Proceedings of the 12th International Workshop on
  Semantic Evaluation. Association for Computational Linguistics, New Orleans,
  Louisiana, pp 39--50, \doi{10.18653/v1/S18-1005},
  \urlprefix\url{https://aclanthology.org/S18-1005}

\bibitem[{Vaswani et~al(2017)Vaswani, Shazeer, Parmar, Uszkoreit, Jones, Gomez,
  Kaiser, and Polosukhin}]{vaswani2017attention}
Vaswani A, Shazeer N, Parmar N, et~al (2017) Attention is all you need. In:
  Advances in neural information processing systems, pp 5998--6008

\bibitem[{Vianna et~al(2023)Vianna, Carneiro, Carvalho, Plastino, and
  Paes}]{vianna2023sentiment}
Vianna D, Carneiro F, Carvalho J, et~al (2023) Sentiment analysis in portuguese
  tweets: an evaluation of diverse word representation models. Language
  Resources and Evaluation pp 1--50

\bibitem[{Wang et~al(2018)Wang, Singh, Michael, Hill, Levy, and
  Bowman}]{wang-2018-glue}
Wang A, Singh A, Michael J, et~al (2018) {GLUE}: A multi-task benchmark and
  analysis platform for natural language understanding. In: Proceedings of the
  2018 {EMNLP} Workshop {B}lackbox{NLP}: Analyzing and Interpreting Neural
  Networks for {NLP}. Association for Computational Linguistics, Brussels,
  Belgium, pp 353--355, \doi{10.18653/v1/W18-5446},
  \urlprefix\url{https://aclanthology.org/W18-5446}

\bibitem[{Wang et~al(2019)Wang, Pruksachatkun, Nangia, Singh, Michael, Hill,
  Levy, and Bowman}]{wang2018superglue}
Wang A, Pruksachatkun Y, Nangia N, et~al (2019) Superglue: A stickier benchmark
  for general-purpose language understanding systems. In: Wallach H, Larochelle
  H, Beygelzimer A, et~al (eds) Advances in Neural Information Processing
  Systems, vol~32. Curran Associates, Inc.,
  \urlprefix\url{https://proceedings.neurips.cc/paper_files/paper/2019/file/4496bf24afe7fab6f046bf4923da8de6-Paper.pdf}

\bibitem[{Wolf et~al(2020)Wolf, Debut, Sanh, Chaumond, Delangue, Moi, Cistac,
  Rault, Louf, Funtowicz, Davison, Shleifer, von Platen, Ma, Jernite, Plu, Xu,
  Le~Scao, Gugger, Drame, Lhoest, and Rush}]{wolf2019huggingface}
Wolf T, Debut L, Sanh V, et~al (2020) Transformers: State-of-the-art natural
  language processing. Proceedings of the 2020 Conference on Empirical Methods
  in Natural Language Processing: System Demonstrations pp 38--45.
  \doi{10.18653/v1/2020.emnlp-demos.6},
  \urlprefix\url{https://aclanthology.org/2020.emnlp-demos.6}

\end{thebibliography}


\begin{appendices}
    
\section{Other considered tasks}

\label{app:other_tasks}

We included other tools and models in \pysentimiento{}. As the tasks are not available in all languages, we have not included them in the main paper. However, we provide a brief description of them in this appendix.

Information extraction is a key task in NLP. In \pysentimiento{}, we provide tools for Named Entity Recognition (NER) and Part-of-Speech (POS) tagging, based on the LinCE corpus \citep{aguilar2020lince}, a large corpus of Spanish-English code-switched tweets. We refer to \cite{perez2022robertuito} for a detailed description benchmark for both tasks in Spanish and English.

In addition to the tasks described in the main paper, we also provide tools for other related tasks or domains, such as contextualized hate speech detection \citep{perez2022contextual} and targeted sentiment analysis \citep{perez2023targeted}, both models specialized in Rioplatense Spanish.




\section{Hyperparameter tuning}

\label{app:training_details}

\begin{table}[h]
    \centering
    \footnotesize
    \begin{tabular}{cc}

        \textbf{Hyperparameter} & \textbf{Values}                          \\
        \hline
        Epochs                  & 3, 4, 5                                  \\
        Batch Size              & 32                                       \\
        Learning Rate           & 2e-5, 3e-5, 5e-5, 6e-5, 7e-5, 8e-5, 1e-4 \\
        Weight Decay            & 0.1                                      \\
        Warmup Ratio            & 0.06, 0.08, 0.10                         \\
        \hline
    \end{tabular}
    \caption{Hyperparameter search space for each considered model.}
    \label{tab:hyperparameters}
\end{table}

To determine the best hyperparameters for each model, we performed a random search using the \textit{wandb} library \citep{wandb}. Table \ref{tab:hyperparameters} summarizes the range of values used for each hyperparameter. We performed between 30 and 60 runs for each model, task, and language, and selected the best model according to the Macro F1 score on the validation set. Following the recommendation of~\cite{tuningplaybookgithub}, we used a batch size of 32, which is the maximum that can fit in our GPU memory (a GTX 1080Ti or Tesla T4, with 11-14GB of memory).

\section{Comparison with other tools}
\label{app:comparison}

\begin{table*}
    \centering
    \footnotesize
    \begin{tabular}{l  l l p{0.25\textwidth} l}
        \hline
        Tool             & Tasks         & Languages                             & Description of algorithm                    & Trained on    \\
        \hline
        \vader{}         & Sentiment     & English                               & Rule and lexicon-based                      & ---           \\
        \textblob{}      & Sentiment     & English                               & Naive Bayes                                 & Movie reviews \\
        \flair{}         & Sentiment     & English                               & DistilBERT                                  &               \\
        \stanza{}        & Sentiment     & English, German                       & BERT?                                       &               \\
        \tweetnlp{}      & Sentiment, HS & Multilingual                          & XLM-T                                       &               \\
        \pysentimiento{} & Sentiment, HS & English, Spanish, Italian, Portuguese & Specialized, monolingual pre-trained models &               \\
        \hline
    \end{tabular}
    \label{tab:compared_tools}
    \caption{Tools compared against \pysentimiento{}.}
\end{table*}

In this section, we describe the comparison with other tools in more detail. We compared \pysentimiento{} with several open-source tools for sentiment analysis and hate speech detection, for different datasets.

\subsection{Open source tools}

\textbf{\vader{}} \citep{hutto2014vader} stands for Valence Aware Dictionary and sEntiment Reasoner. It is a lexicon and rule-based tool specialized in social media data. It provides a sentiment score for each text, which can be positive, negative, or neutral, and also a compound score that represents the overall sentiment of the text ranging from -1 to 1. \textbf{\textblob{}} \citep{loria2018textblob} is an NLP toolkit providing support for common tasks such as NER, POS Tagging, and Sentiment Analysis using classic machine learning techniques. For the task we are concerned with, it provides a polarity Naive Bayes classifier trained on \textit{movie reviews} dataset. Both tools are available for English only.

\textbf{\flair{}} \citep{akbik2019flair} is a general NLP toolkit providing state-of-the-art models for several tasks and languages. For Sentiment Analysis, it provides an English classifier based on the IMDB dataset \citep{maas2011learning}, with DistilBERT \citep{sanh2020distilbert} used as base pre-trained language model.

\textbf{\stanza{}} \citep{qi2020stanza} is another NLP toolkit, built on top of the CoreNLP library. It provides an interface very similar to spaCy, and support for several syntactic and semantic tasks for over 70 languages. For sentiment analysis, it provides models for English, German, and Chinese. Particularly, for the case of English, the model is trained on several datasets, among which the SST-2 dataset is included\footnote{See \url{https://stanfordnlp.github.io/stanza/sentiment.html}}. The exact model used for sentiment analysis is specified neither in the documentation nor in the paper, but according to the code, it seems to be a BERT model\footnote{See \url{https://github.com/stanfordnlp/stanza/blob/main/stanza/utils/training/run_sentiment.py}}.

\textbf{\tweetnlp{}} \citep{camacho-collados2022tweetnlp} is a multilingual NLP toolkit specialized in social media data. It provides models for sentiment analysis and hate speech detection in several languages (among other tasks) based on the XLM-T model \citep{barbieri-2022-xlm}, a fine-tuned version of XLM-R \citep{conneau-2020-unsupervised} on a large dataset of tweets. For the considered tasks,

Table \ref{tab:compared_tools} presents a summary description of the considered tools.

\subsection{Datasets}

\subsubsection*{Sentiment Analysis datasets}

The \textbf{Multilingual Amazon Reviews Corpus} \citep{keung2020marc} is a dataset of Amazon reviews in several languages, including English, Spanish, Chinese, German, and Japanese. While the original work uses MAE to evaluate the performance of the models on a 5-star scale, we converted the scores to positive (4-5 stars), negative (1-2 stars), and neutral (3 stars) to have a common ground with the other corpora.

Another corpus containing several languages is the \textbf{MTEB} Tweet Sentiment Extraction task \cite{muennighoff2023mteb}. While the original paper states the corpus to be one from a Kaggle Competition, we used its multilingual version\footnote{Available at \url{https://huggingface.co/datasets/mteb/tweet_sentiment_multilingual}}, which turns out to be the one provided by the authors of \tweetnlp{} \citep{barbieri-2022-xlm}. This dataset is available in Arabic, English, French, German, Hindi, Italian, Portuguese and Spanish. The tags are positive, negative, and neutral.

The \textbf{Stanford Sentiment Treebank} \citep{socher2013recursive} provides movie reviews with fine-grained sentiment labels, set at different levels of the parse tree. The binary version of the dataset, SST-2, is used in this comparison, which assigns a positive/negative label to the whole sentence. The \textbf{SentEval} dataset \citep{conneau2018senteval} is a benchmark for evaluating the quality of sentence embeddings. Among its tasks, it includes Sentiment Analysis. We selected the Customer Reviews task \citep{hu2004mining}. These datasets are available exclusively in English, and both have a positive/negative classification task associated.

The \textbf{Financial Phrasebank} \citep{malo2014financial} is a dataset of financial news headlines annotated with sentiment labels. The \textbf{Sentiment140} dataset \citep{go2009twitter} is a classical corpus of tweets annotated for this task. Both datasets are in English and share the positive/negative/neutral classification task.

For Italian, we also used the \textbf{FEEL-it} dataset \citep{bianchi2021feel}. This same dataset was used to train the \pysentimiento{} model for emotion detection.

For all these corpora, we selected the test split whenever available. In the case of the SST-2, we used the evaluation set.

\subsubsection*{Hate Speech Detection datasets}

The \textbf{HateCheck} dataset \cite{rottger2021hatecheck} is a dataset of synthetic hate speech examples, created by first eliciting desirable ``functionalities'' of a hate speech classifier, and then by generating templates addressing these requirements. The dataset is available in English, and we also used translated versions of it for Spanish, Italian, and Portuguese provided by the main author\footnote{\url{https://huggingface.co/datasets/Paul/hatecheck-spanish}}. The classification task is binary, with the labels being hateful vs non-hateful.

\end{appendices}

\end{document}